\documentclass{article}

\usepackage{arxiv}

\usepackage[utf8]{inputenc} 
\usepackage[T1]{fontenc}    
\usepackage{hyperref}       
\usepackage{url}            
\usepackage{booktabs}       
\usepackage{amsfonts}       
\usepackage{nicefrac}       
\usepackage{microtype}      
\usepackage{lipsum}		
\usepackage{graphicx}
\usepackage{natbib}
\usepackage{doi}
\usepackage{amsmath}
\usepackage{float}
\usepackage{multirow}

\title{GuideCAD: A lightweight multimodal framework for 3D CAD model generation via prefix embedding}

\author{
    Minseong Kim\\
    Convergence Research Center for Insect Vectors\\
    Incheon, South Korea\\
    \texttt{qtddpms@gmail.com}
    \And
    Jinyeong Park\\
    Department of Computer Science and Engineering\\
    Incheon National University\\
    Incheon, 22012, South Korea
    \And
    Sungho Park\thanks{Corresponding authors: Sungho Park (yunisomi@inu.ac.kr) and Jibum Kim (jibumkim@inu.ac.kr).}\\
    Department of Computer Science and Engineering\\
    Incheon National University\\
    Incheon, 22012, South Korea\\
    \texttt{yunisomi@inu.ac.kr}
    \And
    Jibum Kim\footnotemark[1]\\
    Department of Computer Science and Engineering\\
    Center for Brain-Machine Interface\\
    Incheon National University\\
    Incheon, South Korea\\
    \texttt{jibumkim@inu.ac.kr}
}

\renewcommand{\shorttitle}{\textit{arXiv} Template}

\hypersetup{
pdftitle={A template for the arxiv style},
pdfsubject={q-bio.NC, q-bio.QM},
pdfauthor={David S.~Hippocampus, Elias D.~Striatum},
pdfkeywords={First keyword, Second keyword, More},
}

\newcommand{\curly}[1]{\{#1\}}

\begin{document}
\maketitle

\begingroup
\renewcommand\thefootnote{}
\footnotetext{
This manuscript is an author-accepted version of the article published in IEEE Access.
The final version is available at https://doi.org/10.1109/ACCESS.2025.3604810.
}
\endgroup

\begin{abstract}
Multi-modal approaches used for 3D CAD generation require substantial computational resources, necessitating efficient training. To address this, we propose GuideCAD, which leverages semantically rich visual-textual representations having only a small number of trainable parameters to generate 3D CAD models. Specifically, GuideCAD uses a mapping network that converts image embeddings into prefix embeddings, enabling a pretrained large language model to integrate visual and textual information. As a result, a transformer-based decoder predicts the construction sequence using the visual-textual embeddings in order to generate the 3D CAD model. For experimental evaluation, we construct a new dataset, referred to as GuideCAD, which consists of text-image pairs. Each pair includes a text prompt that represents a 3D CAD construction sequence and its corresponding 3D CAD image. Our experimental results show that GuideCAD generates comparably high-quality 3D CAD models while using approximately four times fewer parameters and achieving twice the training efficiency compared to fine-tuning approaches. We have released the source code and dataset for our method at: https://github.com/mskimS2/GuideCAD.
\end{abstract}


\section{Introduction}

\begin{figure*}[!t]
\centering    
\includegraphics[width=0.85\textwidth,height=6cm]{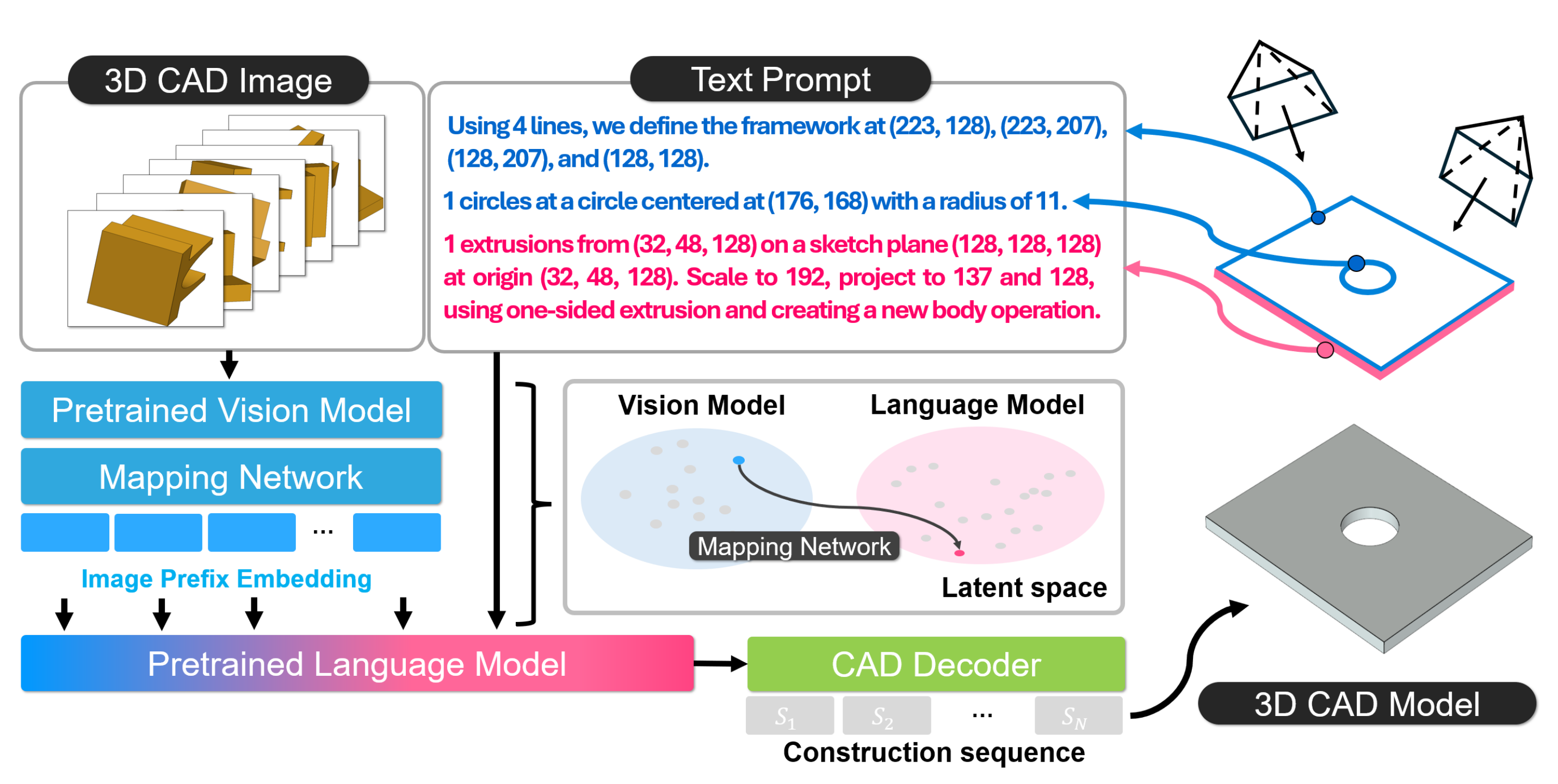}
\caption{GuideCAD efficiently generates 3D CAD models from a text prompt and a corresponding 3D CAD image. A language model integrates image and text embeddings into a visual-textual representation, which guides the decoder in predicting the construction sequence for 3D CAD models.}
\label{fig:fig1}
\end{figure*}

CAD is an essential tool across various domains such as product design, circuit design, and parts manufacturing. CAD facilitates the visualization of ideas and their application in the design workflow through the process of 3D modeling. Professional designers often create 3D CAD models using commercial software (e.g., SolidWorks \cite{solidworks}, FreeCAD \cite{freecad}, and Para-Solid \cite{parasolid}), which are then used to optimize simulation and manufacturing processes. 3D CAD model can be constructed through a sequence of operations known as a construction sequence ($S$), typically composed of commands such as \textit{Sketch} and \textit{Extrusion}. The resulting model can be stored either as the construction sequence itself or in formats such as boundary representation (B-Rep). This workflow provides flexibility by allowing designers to track the design history of the 3D CAD model and make iterative modifications as needed.

Generating a 3D CAD model is a challenge that requires expert knowledge, as it involves not only creating geometric shapes but also considering functional elements \cite{CAMBA201618}. Recently, to address these challenges, there has been research on generating 3D CAD models by learning from text descriptions of the desired geometric and functional properties of the models \cite{text2cad,CAD-MLLM,Generating_CAD_Code}. Text-based approaches allow users to easily describe and generate the desired 3D CAD models using only text. In addition, image-based methods for generating 3D CAD models have also been explored \cite{CAD-MLLM,Generating_CAD_Code,img2cad,query2cad,OpenECAD,gencad}. These image-based approaches are particularly effective at capturing the structural features and overall shapes of 3D CAD models that are difficult to describe using text descriptions. 

However, such methods that rely only on a single modality have limited ability to fully represent the complex structural and functional characteristics of 3D CAD models. For instance, text-only approaches may find it challenging to consistently represent the detailed structures of 3D CAD models, as these structures can vary depending on the viewing angle \cite{text2cad}. In response to these challenges, recent studies have focused on generating 3D CAD models by fine-tuning pretrained large language models (LLMs) using multi-modal inputs \cite{CAD-MLLM,OpenECAD,rukhovich2024cadrecode}. Fine-tuning approaches based on multi-modal data can generate more refined and complicated 3D CAD models by complementing the incomplete information present in each individual modality \cite{CAD-MLLM,multicad}. However, the large number of trainable parameters in LLMs results in higher computational cost and training time \cite{pept}.

To overcome this issue, we propose GuideCAD, a novel framework that integrates multi-modal data (\textit{i.e.,} image and text prompt) based on prefix embeddings, thereby enhancing the quality of 3D CAD models with efficient computational cost. As illustrated in Figure \ref{fig:fig1}, given a 3D CAD image and a corresponding text prompt describing its construction sequence, it transforms an image embedding extracted by the pretrained vision model into an image prefix embedding ($P$) using a mapping network. By incorporating this image prefix embedding into the input space of the subsequent LLM, the multi-modal information is seamlessly integrated into a unified visual–textual representation, which is then used by the transformer-based CAD decoder \cite{transformer} to generate a construction sequence. Following the philosophy of prompt tuning \cite{pept,jia2022vpt}, we freeze the entire pretrained encoder in this process and introduce only a small number of trainable parameters to ensure training efficiency. Specifically, only additional parameters for the mapping network and generating the construction sequence are required. This approach not only enables flexible integration of various modalities, but also allows for efficient model training with substantially lower computational resources.

To empirically validate the effectiveness of the proposed method, we construct a new dataset, referred to as GuideCAD, consisting of text prompts and 3D CAD image pairs. Building upon the existing benchmark (\textit{i.e., }the DeepCAD Dataset \cite{deepcad}), we perform three key data generation and preprocessing steps:
(1) Generation of construction sequences from 3D CAD models,
(2) Generation of text prompts describing the construction sequences using a text template guided by an LLM, and (3) Creation of multi-view image data by capturing 3D CAD models from various angles using pythonOCC \cite{pythonocc}. 
Experimental results show that GuideCAD significantly outperforms state-of-the-art vision-language models (VLMs) and achieves generation quality comparable to that of fine-tuning approaches, while remarkably reducing the number of trainable parameters by a factor of four and cutting total training time by more than half. Furthermore, it successfully predicts accurate construction sequences, producing high-quality 3D CAD models that are close to the ground truth. 

Our contributions can be summarized as follows:
\begin{itemize}
\item We propose GuideCAD, a lightweight framework that learns only a small number of parameters to generate 3D CAD models from multi-modal inputs consisting of text prompts and 3D CAD images. \
\item We introduce a mapping network to efficiently integrate visual-textual representation for learning and generation of 3D CAD models. \
\item GuideCAD demonstrates superior performance to state-of-the-art vision-language models in generating high-fidelity 3D CAD models that reflect the intended design more accurately.
\item We release the GuideCAD dataset, a publicly available multi-modal dataset that pairs multi-view images of 3D CAD models with corresponding text prompts.
\end{itemize}

\section{Related works}
\label{sec:related works}
\subsection{Construction Language Modeling for 3D Shape}
As the automation of 3D CAD modeling has been increasing, researchers are focusing on conceptualizing the CAD design process as language modeling. \cite{cadlanguage,Sketchgen} generate sketches of 2D CAD models using token sequences to represent complex constraints. To generate 3D CAD models, many approaches use a parameterized construction sequence consisting of \textit{Sketch} and \textit{Extrusion} \cite{deepcad,skexgen,cadparser,contrastcad,Brep2Seq}. They learn to predict command and parameter tokens from each parameterized construction sequence. DeepCAD \cite{deepcad} reconstructs 3D CAD models using the ABC dataset \cite{Koch_2019_CVPR}, which represents construction sequences based on \textit{Sketch} and \textit{Extrusion}. Another approach to 3D modeling involves representing objects using relationships between vertices, edges, and faces. Polygen \cite{polygen} combines Transformer \cite{transformer} and Pointer Networks \cite{pointernetwork} to sequentially predict the vertices and faces of a 3D Polygon Mesh. Similarly, TurtleGen \cite{turtlegen} improves by representing 3D CAD models as sketch graphs, allowing the learning of relationships between vertices and edges. ComplexGen \cite{complexgen} and BrepGen \cite{brepGen} reconstruct 3D CAD models using Boundary Representation (B-Reps). Researchers have also conducted studies on generating 3D CAD models using multi-stage learning strategies \cite{deepcad,skexgen,hnc,multicad,CAD-SIGNet,drawstepbystep}. \cite{hnc} propose a novel generative model for controllable CAD design, leveraging a three-level neural code to capture design patterns and intent. In contrast, GuideCAD learns to predict the construction sequence using a single-stage strategy.

\subsection{Diverse Input Representations for 3D CAD generation}
For easier CAD modeling, researchers have used various input modalities, such as images, texts, and point clouds. Among them, text is considered one of the simplest approaches for generating 3D CAD models \cite{text2cad,cadtranslator,mechnicalllm}. In Text2CAD \cite{text2cad}, a text prompt generated with a visual language model is used as a conditional input to a decoder consisting of a transformer for 3D CAD generation. In contrast, some works focus on generating 3D CAD models conditioned on images \cite{img2cad,query2cad,OpenECAD,gencad} or on the drawing processes of CAD designers \cite{CAD2Program,free2cad,vitruvion}. For 3D CAD model generation via reverse engineering, Img2CAD \cite{img2cad} employs images as input, while \cite{voxel2cad} uses point clouds. MultiCAD \cite{multicad} leverages multi-modal contrastive learning to integrate geometric features with the construction sequences. Following these trends, several approaches have attempted to integrate and leverage multi-modalities. \cite{point-e,rukhovich2024cadrecode} incorporate point clouds and text, whereas \cite{CAD-MLLM,Generating_CAD_Code,OpenECAD,cadvlm,cadgptsynthesising} rely on text-image pairs to guide the generation of 3D CAD models. Similarly, our approach efficiently generates 3D CAD models by learning from text-image inputs through additional tuning of model parameters.

\subsection{Effectiveness of Lightweight Tuning}
Some works focus on efficiently adapting large pretrained models by updating only the parameters of a minimal module, enabling flexible adaptation to specific tasks. \cite{prefixtuning,jia2022vpt,liu-etal-2022-p,hu2022lora} use fewer learnable parameters within pretrained models, allowing for effective adaptation to specific tasks. \cite{pept} use prompt tuning for model training, achieving performance comparable to full fine-tuning. To efficiently adapt large pretrained models, \cite{clipcap,sia-duh-2022-prefix,gao2023prefixmol} focus on training prefix embeddings, allowing the models to perform various tasks while preserving the performance of pretrained models. The advantage of these approaches is that they save time and computational cost compared to fine-tuning. We explore the use of prefix embeddings to enhance the relationship between visual and textual representations, focusing on predicting the construction sequence for 3D CAD model generation. 

\begin{figure*}[!t]
\centering    
\includegraphics[width=\textwidth,height=7.5cm]{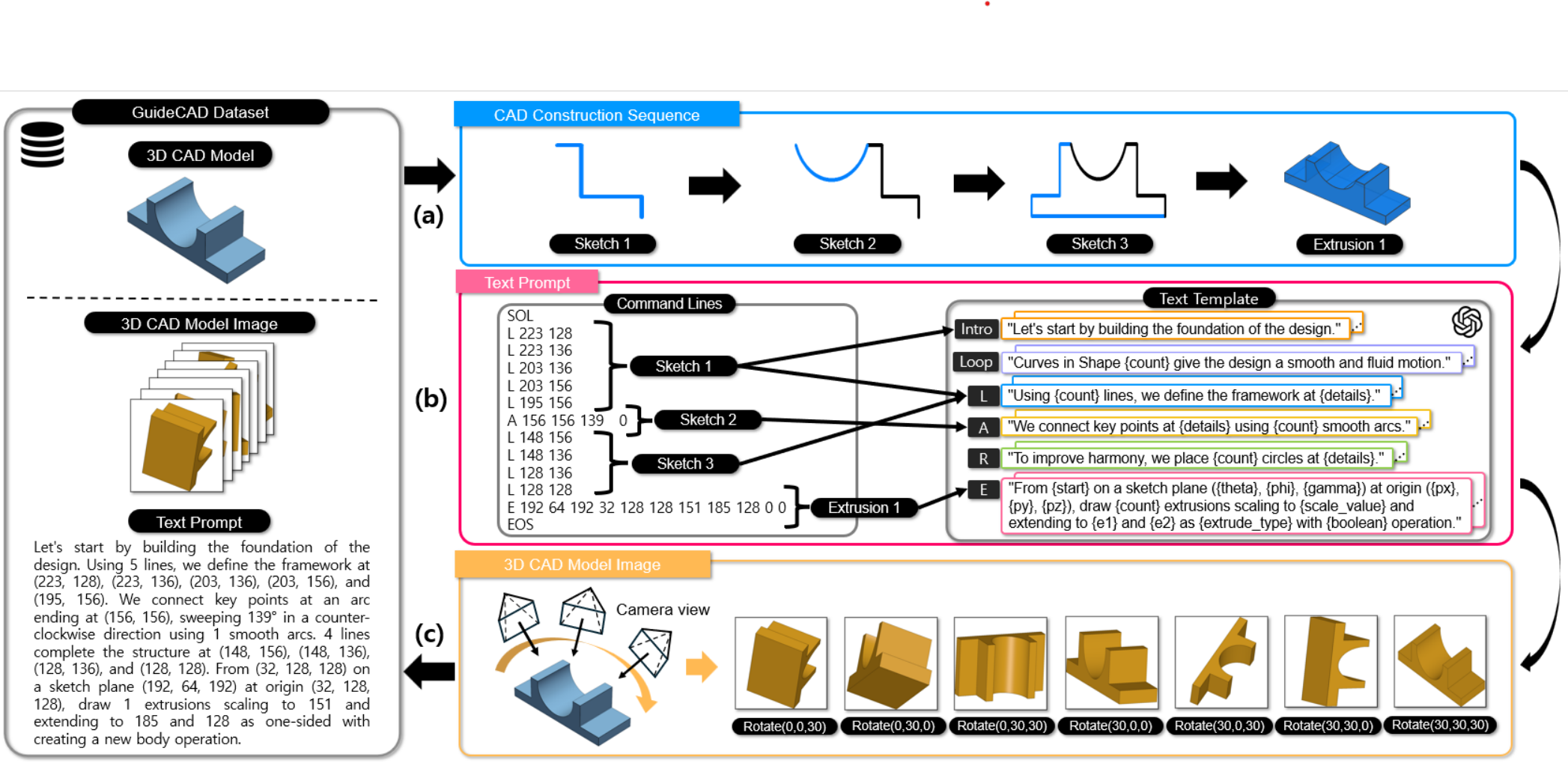}
\caption{Three-step process for generating the GuideCAD dataset. (a) Step 1: Generation of construction sequence, consisting of \textit{Sketch} and \textit{Extrusion} commands, from a 3D CAD model (b) Step 2: Creating the text prompt from a text template, generated by the LLM (c) Step 3: Capturing multi-view images from the rendered 3D CAD model using pythonOCC.}
\label{fig:figure2}
\end{figure*}

\section{Construction Sequence-Based Annotation for GuideCAD Dataset}
While the construction sequence offers a detailed, step-by-step representation of the 3D CAD modeling process, modifying it to achieve a specific target shape requires expert knowledge to correctly align the sequence with the intended design. To address this, our approach adopts text prompts and 3D CAD images as input data, providing users with a more intuitive alternative to the construction sequence. In previous works, the common approach of using only text prompts can lead to inconsistent representations of the same 3D CAD model, making it difficult to express specific structural details\cite{text2cad}. Therefore, we integrate images as well as text prompts into the input space, enabling a direct representation of shape features and consequently facilitating a more precise understanding of the structure of CAD model. Nevertheless, given that there is currently no publicly available dataset containing paired 3D CAD images, text prompts, and construction sequences, we start by constructing the GuideCAD dataset through a three-step process, generating the construction sequence, text prompt, and 3D CAD image independently. 

\subsection{CAD Construction Sequence Generation}
\begin{table}[]
\centering
\caption{Commands and corresponding parameters in the construction sequence for 3D CAD model generation.}
\begin{tabular}{c c l}
\hline
\textbf{Command} & \textbf{Parameter} & \textbf{Description} \\ \hline
<SOL> & $\emptyset$ & start of line \\ \hline
L (Line) & $x$, $y$ & line end point point ($x$, $y$) \\ \hline
\multirow{3}{*}{A (Arc)} & $x$, $y$ & arc end point ($x$, $y$) \\ \cline{2-3} 
& $\theta$ & angle of the arc \\ \cline{2-3} 
& $c$ & counter-clockwise flag \\ \hline
\multirow{2}{*}{R (Circle)} & $x$, $y$ & circle center point ($x$, $y$) \\ \cline{2-3} 
& $r$ & circle radius \\ \hline
\multirow{6}{*}{E (Extrusion)} & $p_x$, $p_y$, $p_z$ & orientation of sketch plane \\ \cline{2-3} 
& $o_x$, $o_y$, $o_z$ & origin of sketch plane \\ \cline{2-3} 
& $s$ & scale of the sketch profile \\ \cline{2-3} 
& $e_1$, $e_2$ & extrude both sides \\ \cline{2-3} 
& $b$ & merge type \\ \cline{2-3} 
& $w$ & extrude type \\ \hline
EOS & $\emptyset$ & end of line \\ \hline
\end{tabular}%
\label{tab:cadsequence}
\end{table}

In the first step, we represent various 3D CAD models from the DeepCAD \cite{deepcad} dataset as construction sequences, consisting of \textit{Sketch} and \textit{Extrusion} commands. Figure \ref{fig:figure2} (a) illustrates the process of generating a 3D CAD model from a construction sequence. The \textit{Sketch} creates closed curves on a 2D plane, with each loop beginning with the <SOL> command. Following the <SOL> commands, three sequential commands are listed: Line (L), Arc (A), and Circle (R). The \textit{Extrusion} extends the 2D shape defined by the 2D Sketch and includes parameters such as orientation, origin, extrusion type (e.g., one-sided, symmetric, two-sided), merge type (e.g., new body, joining, cutting, intersecting), sketch profile scale, and extrusion distance. Each construction sequence ends with the <EOS> command. In this study, we use a total of $N_c=6$ commands, which include \{<SOL>, Line, Arc, Circle, Extrusion, <EOS>\} and $N_p=16$ parameters, including \{$x$, $y$, $\theta$, $c$, $r$, $p_x$, $p_y$, $p_z$, $o_x$, $o_y$, $o_z$, $s$, $e_1$, $e_2$, $b$, $w$\}, as summarized in Table~\ref{tab:cadsequence}.

\subsection{Text Prompt Generation using an LLM}
\begin{table*}[t!]
\centering
\caption{We use GPT-4 \cite{gpt2} to generate text templates, where each template represents the corresponding construction sequence parameters. The symbol "\{\}" serves as a placeholder in the text template, which is filled with appropriate parameter values. The \{count\} denotes the number of consecutive occurrences of a specific command in the construction sequence.}
\resizebox{\textwidth}{!}{%
    \begin{tabular}{c l}
    \hline
    \textbf{Command}   & \textbf{Text Template} \\ \hline
    Introduce & 
    \begin{tabular}[c]{@{}l@{}}
    - Let's start by building the foundation of the design.\\
    - We begin by sketching the basic structure.\\
    - Starting with key shapes helps define the layout.\\
    - The design starts to take form with simple elements.\\
    - We lay out the initial framework to guide the model.\\
    - Sketching lines and curves gives the model its foundation.\\
    - This step establishes the symmetry and balance of the design.\\
    - We start shaping the model with basic geometric features.\\
    - Building a clear framework sets the tone for the design.\\
    - The design begins to take shape with well-placed elements.
    \end{tabular} \\ \hline
    
    L (Line)  & 
    \begin{tabular}[c]{@{}l@{}}
    - Let's connect \curly{count} lines to form a framework at \curly{details}.\\
    - Using \curly{count} lines, we define the framework at \curly{details}.\\
    - We use \curly{count} lines to connect key points like \curly{details}.\\
    - With \curly{count} lines, we form the edges at \curly{details}.\\
    - We use \curly{count} lines to make smooth transitions in this stage.\\
    - The structure comes together with \curly{count} lines at \curly{details}.\\
    - \curly{count} lines complete the structure at \curly{details}.
    \end{tabular} \\ \hline
    
    A (Arc) & 
    \begin{tabular}[c]{@{}l@{}}
    - We connect key points at \curly{details} using \curly{count} smooth arcs.\\
    - Let's draw \curly{count} arcs to create fluid curves around \curly{details}.\\
    - Using \curly{count} arcs, we add elegance to the layout at \curly{details}.\\
    - We shape the structure with \curly{count} gentle arcs near \curly{details}.\\
    - Drawing \curly{count} arcs at \curly{details} helps refine the design's flow.\\
    - With \curly{count} arcs, we ensure a smooth transition at \curly{details}.\\
    - We add {count} graceful arcs to guide the design near {details}.\\
    - \curly{count} arcs flow naturally around \curly{details}, enhancing the layout.\\
    - Curves made with \curly{count} arcs bring balance to {details}.\\
    - Let's enhance the geometry by sketching \curly{count} arcs at \curly{details}.
    \end{tabular} \\ \hline
    
    R (Circle) & 
    \begin{tabular}[c]{@{}l@{}}
    - We add \curly{count} circular shapes near \curly{details} to enhance symmetry.\\
    - Drawing \curly{count} circles around \curly{details} helps balance the layout.\\
    - Let's position \curly{count} circles at \curly{details} to refine the proportions.\\
    - {count} circular features near \curly{details} bring structure and clarity.\\
    - To improve harmony, we place \curly{count} circles at \curly{details}.\\
    - Adding \curly{count} precise circles at {details} anchors the design.\\
    - We create smooth transitions by adding \curly{count} circles around \curly{details}.\\
    - The layout is enhanced with \curly{count} circular elements near \curly{details}.\\
    - Positioning \curly{count} circles at \curly{details} ensures a balanced structure.\\
    - The geometry gains clarity with \curly{count} circles placed near \curly{details}.
    \end{tabular} \\ \hline
    
    E (Extrusion) & 
    \begin{tabular}[c]{@{}l@{}}
    - Start at ({$p_x$}, {$p_y$}, {$p_z$}) on a sketch plane oriented ({$\theta$}, {$\phi$}, {$\gamma$}) with origin at ({$p_x$}, {$p_y$}, {$p_z$}). Create \curly{count} extrusions scaled by {$s$}, extending to {$e_1$} and {$e_2$} with {$w$} extrusion and {$b$} operation.\\
    - From ({$p_x$}, {$p_y$}, {$p_z$}) on a sketch plane ({$\theta$}, {$\phi$}, {$\gamma$}) at origin ({$p_x$}, {$p_y$}, {$p_z$}), draw \curly{count} extrusions scaling to {$s$} and extending to {$e_1$} and {$e_2$} as {$w$} with {$b$} operation.\\
    - Create \curly{count} extrusions at ({$p_x$}, {$p_y$}, {$p_z$}) on a sketch plane ({$\theta$}, {$\phi$}, {$\gamma$}) with origin ({$p_x$}, {$p_y$}, {$p_z$}), scaling by {$s$} to reach {$e_1$} and {$e_2$}. Extrusion: {$w$}, Operation: {$b$}.\\
    - Draw \curly{count} extrusions starting at ({$p_x$}, {$p_y$}, {$p_z$}) on a sketch plane ({$\theta$}, {$\phi$}, {$\gamma$}) with origin at ({$p_x$}, {$p_y$}, {$p_z$}). Scale {$s$}, extend to {$e_1$} and {$e_2$}, and use {$w$} extrusion with {$b$} operation.\\
    - Begin \curly{count} extrusions from ({$p_x$}, {$p_y$}, {$p_z$}) on a sketch plane ({$\theta$}, {$\phi$}, {$\gamma$}) at origin ({$p_x$}, {$p_y$}, {$p_z$}). Scale to {$s$}, project to {$e_1$} and {$e_2$}, using {$w$} extrusion and {$b$} operation.\\
    - \curly{count} extrusions start at ({$p_x$}, {$p_y$}, {$p_z$}), on a sketch plane ({$\theta$}, {$\phi$}, {$\gamma$}), with origin ({$p_x$}, {$p_y$}, {$p_z$}). Scale: {$s$}, Extend: {$e_1$}, {$e_2$}, Type: {$w$}, Operation: {$b$}.\\
    - Add \curly{count} extrusions at ({$p_x$}, {$p_y$}, {$p_z$}) on a sketch plane ({$\theta$}, {$\phi$}, {$\gamma$}) with origin ({$p_x$}, {$p_y$}, {$p_z$}). Scale by {$s$}, extend to {$e_1$} and {$e_2$}, using {$w$} type and {$b$} operation.\\
    - Create \curly{count} extrusions starting from ({$p_x$}, {$p_y$}, {$p_z$}), on a sketch plane ({$\theta$}, {$\phi$}, {$\gamma$}), with origin ({$p_x$}, {$p_y$}, {$p_z$}). Scale {$s$}, extend to {$e_1$} and {$e_2$}, and apply {$w$} extrusion with {$b$} operation.\\
    - At ({$p_x$}, {$p_y$}, {$p_z$}) on a sketch plane ({$\theta$}, {$\phi$}, {$\gamma$}) with origin ({$p_x$}, {$p_y$}, {$p_z$}), draw \curly{count} extrusions scaled by {$s$} to reach {$e_1$} and {$e_2$}. Extrusion type: {$w$}, Boolean operation: {$b$}.\\
    - Draw \curly{count} extrusions starting from ({$p_x$}, {$p_y$}, {$p_z$}) on a sketch plane ({$\theta$}, {$\phi$}, {$\gamma$}) at origin ({$p_x$}, {$p_y$}, {$p_z$}), scaled to {$s$}, extending to {$e_1$} and {$e_2$}, with {$w$} type and {$b$} operation.
    \end{tabular} \\ \hline
    
    Loop & 
    \begin{tabular}[c]{@{}l@{}}
    - Shape \curly{count} brings together curves and edges to balance the design.\\
    - In Shape \curly{count}, we refine the transitions to make the structure smooth.\\
    - Shape \curly{count} improves the design by combining arcs and lines.\\
    - This stage in Shape \curly{count} focuses on symmetry and alignment for a clean look.\\
    - In Shape \curly{count}, soft curves and sharp lines come together seamlessly.\\
    - Paths in Shape \curly{count} are adjusted to create better flow and balance.\\
    - Shape \curly{count} enhances the structure by blending arcs and straight lines.\\
    - This phase in Shape \curly{count} adds clarity with carefully defined paths.\\
    - Curves in Shape \curly{count} give the design a smooth and fluid motion.\\
    - Shape \curly{count} aligns key elements for a polished and cohesive layout.
    \end{tabular} \\ \hline

    Details & 
    \begin{tabular}[c]{@{}l@{}}
    - L (Line): ({$x_1$},{$y_1$}), ({$x_2$}, {$y_2$}), $...$, and ({$x_n$}, {$y_n$}) \\
    - A (Arc): an arc ending at ({$x$}, {$y$}), sweeping {$\theta$} ° in a {$c$} direction \\
    - R (Circle): a circle centered at ({$x$}, {$y$}) with a radius of {$r$}
    \end{tabular} \\ \hline
    
    \end{tabular} 
}
\label{tab:prompt-template}
\end{table*}
Next, the pretrained LLM (\textit{i.e.,} GPT-4 \cite{gpt4}) generates a text template used to create the text prompt, which consists of the command and parameter values from the construction sequence. Table~\ref{tab:prompt-template} presents the generated text prompt templates used in this study. The templates, each representing specific commands (e.g., L, A, R, E), are designed to clearly describe the commands along with their corresponding parameters. Figure \ref{fig:figure2} (b) shows the Command Lines, where each command and its corresponding parameter from the construction sequence are listed on separate lines. Subsequently, as shown in Figure \ref{fig:figure2} (b), the listed Command Lines are converted into text prompts using a pre-generated template, and these prompts are then combined to form the complete prompt. 

\subsection{Multi-view 3D CAD Image Generation}
Finally, multi-view images are generated for the given 3D CAD model rendering it using the construction sequence with pythonOCC \cite{pythonocc}. As a single-view image of the 3D CAD model could obscure certain features, making it difficult to capture the complete shape, a multi-view image from different angles is necessary for a more accurate representation\cite{xie2019pix2vox, 8453803}. As illustrated in Figure \ref{fig:figure2} (c), a total of seven images were captured by rotating the model by 30 degrees along the XYZ axes at the following angles: 
$(0, 0, \frac{\pi}{6})$, 
$(0, \frac{\pi}{6}, 0)$, 
$(0, \frac{\pi}{6}, \frac{\pi}{6})$, 
$\left(\frac{\pi}{6}, 0, 0\right)$,
$\left(\frac{\pi}{6}, 0, \frac{\pi}{6}\right)$,
$\left(\frac{\pi}{6}, \frac{\pi}{6}, 0\right)$, 
$\left(\frac{\pi}{6}, \frac{\pi}{6}, \frac{\pi}{6}\right)$. 
Consequently, the method captures the shape of the CAD model efficiently by reducing occlusions and using only the necessary viewpoints. To handle the varying sizes and shapes of 3D CAD models, the camera position is adjusted to avoid cropping or obscuring any part of the CAD model, ensuring it remains centered in the image at $1024\times768$ resolution. 


\section{GuideCAD Architecture}

\label{sec:GuideCAD Architecture}

\begin{figure*}[!t]
\centering
\includegraphics[width=0.85\textwidth,height=0.23\textheight]{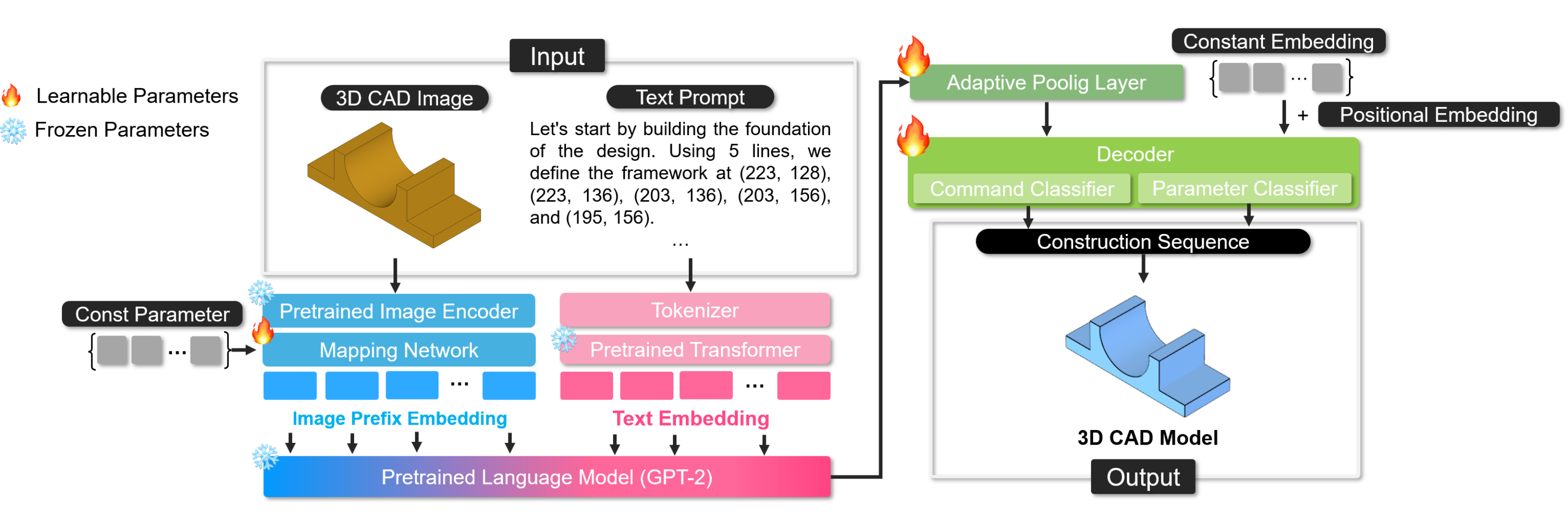}
\caption{Overview of the proposed GuideCAD architecture. During training, GuideCAD updates only the Mapping Network and the modules required for predicting the construction sequence, while keeping all pretrained model parameters frozen.}
\label{fig:figure3}
\end{figure*}

Figure \ref{fig:figure3} provides an overview of GuideCAD, which aims to generate 3D CAD models from image-text prompts $\{I^i, T^i\}_{i=1}^N$, where $N$ represents the total number of pairs. Each pair consists of a 3D CAD image $I^i \in \mathbb{R}^{H \times W \times 3}$ with height $H$, width $W$, and three RGB channels, along with its corresponding tokenized text prompt $T^{i}$.

A randomly selected image, rotated along the XYZ axes from a set of seven generated images, is paired with the corresponding text prompt. The selected image is processed using the pretrained image encoder ($f_{enc}$) to generate image embeddings, which are then transformed into Prefix Embeddings ($P$) via the Mapping Network ($f_m$). The $f_m$ converts the image embedding into a prefix form that combines with the text embedding, allowing the language model to learn both modalities. The transformed $P$ is used as the prefix embedding in the pretrained language model. Following the Adaptive Pooling Layer, the final layer embeddings of the language model are used by the Decoder to predict the construction sequence. The Adaptive Pooling Layer assists the Decoder to predict an accurate construction sequence by adjusting the final layer embedding of language model to CAD-specific visual and textual features.

One of the main differences in GuideCAD is the incorporation of a prefix embedding that includes visual information before the text tokens in the input space. By freezing the weights of the pretrained LLM and training only the Mapping Network for generating the prefix embeddings, the model not only captures a joint visual-textual representation but also significantly reduces the number of learnable parameters. To the best of our knowledge, due to the lack of large-scale pretrained models specialized for the CAD domain, we use CLIP \cite{clip}—which is pretrained on large-scale image-text pairs with strong cross-modal alignment—and GPT-2 \cite{gpt2} as the image encoder and language model, respectively. 

\subsection{Mapping Network}
The Mapping Network ($f_m$) uses shallow transformer layers to transform the image embedding ($E_I$ ), obtained through $f_{enc}$, into the image prefix embedding $P$. First, the input 3D CAD Image ($I$) is converted into the image embedding $E_I$ through $f_{enc}$. This process is defined as follows:
\begin{equation}
E_I=f_{enc} (I),
\end{equation}
where $E_I \in \mathbb{R}^{d_{clip}}$, $d_{clip}$ represents the dimensionality of the image embedding obtained through the CLIP image encoder \cite{clip}.

To incorporate the visual representation into the input space of the pretrained language model, $f_m$ converts $E_I$ into the prefix embedding $P$, defined as \( f_m: \mathbb{R}^{d_{\text{clip}}} \to \mathbb{R}^{p_l \times 768} \), where $p_l$ denotes the prefix length. We introduce a learnable constant parameter $C = \{ C_1, C_2, C_3, \dots, C_{p_c} \} \in \mathbb{R}^{p_c \times 768}$, similar to the approaches in \cite{prefixtuning, jia2022vpt} to model the relationship between $C$ and $E_I$, enhancing the strong representational power of $P$. Each $C_i \in \mathbb{R}^{768}$ is a learnable parameter, defined based on the $p_c$, and treated as a learnable token that is concatenated with $E_I$ before being passed into the model. To combine $E_I$ with $C$, we apply a linear transformation to project $E_I$ into a $p_l \times 768$ embedding space, which is defined as follows:
\begin{equation}
E_{proj}=W_{proj}E_I,
\end{equation}
where $W_{proj} \in \mathbb{R}^{(p_l \times 768) \times d_{\text{clip}}}$ is a learnable weight that projects $E_I$ from the $d_{\text{clip}}$ dimension to the $p_l \times 768$ dimensional embedding space, and $E_{\text{proj}} \in \mathbb{R}^{p_l \times 768}$. 

Subsequently, $f_m$ generates $P$ by concatenating $E_{\text{proj}}$ with the constant parameter $C$. This process is defined as follows:
\begin{equation}
P=f_m([C;E_{proj}]),
\end{equation}
Here, $[C; E_{proj}] \in \mathbb{R}^{(p_c + p_l) \times 768}$ represents the concatenation of the learnable constant parameter $C$ and $E_{proj}$, and $P \in \mathbb{R}^{(p_c+p_l) \times 768}$. 

To simplify the model, we set $p_c$ and $p_l$ 
to the same value, enabling $f_m$ to learn both embeddings in a consistent manner. We then obtain $P \in \mathbb{R}^{p_l \times 768}$ by selecting only the part containing the visual information. 

\subsection{Pretrained Language Model (GPT-2)}
To generate the visual-textual embedding, we combine the text embedding with $P$ and subsequently encode it using the pretrained GPT-2 model \cite{gpt2}. Specifically, the text tokens $T$ are converted into a text embedding $E_T \in \mathbb{R}^{N_t \times 768}$, where $N_t$ represents the length of $T$. As given in Equation (\ref{gpt2_equation}), the $P$ and $E_T$ are concatenated and provided to the pretrained GPT-2 model, which extracts embeddings from the final layer to produce $ z_{gpt} \in \mathbb{R}^{(p_l + N_t) \times 768}$. The resulting $z_{gpt}$ represents a semantically rich visual-textual representation.
\begin{equation}
z_{gpt}=\text{GPT-2}([P;E_T]),
\label{gpt2_equation}
\end{equation}
where $[P; E_T]$ denotes the concatenation of $P$ and $E_T$, such that $[P; E_T] \in \mathbb{R}^{(p_l + N_t) \times 768}$.

\subsection{Adaptive Pooling Layer}
An adaptive pooling layer refines $z_{gpt}$ to align with visual-textual representations specific to the CAD modeling tasks, facilitating the generation of 3D CAD models. In Equation (\ref{apl}),  the $z_{gpt}$ is passed through the Adaptive Pooling Layer (APL), producing the latent embedding ($z$).
\begin{equation}
z=\text{AdaptivePoolingLayer}(z_{gpt}) (6),
\label{apl}
\end{equation}
where $z \in \mathbb{R}^{E_a \times 768}$, with $E_a$  representing the adjusted token length. 

\subsection{Decoder}
Similar to DeepCAD \cite{deepcad}, the decoder focuses on predicting the construction sequence to generate 3D CAD model by passing constant embedding ($E_c \in \mathbb{R}^{N_s\times768}$) and $z$. To address this, we learn the probability distribution $\phi$, defined as follows: 
\begin{equation}
\phi(\hat{S} \mid z, E_c) = \prod_{i=1}^{N_s} \phi(\hat{S}_{c_i}, \hat{S}_{p_i} \mid z, E_c, \Theta),
\end{equation}
where $\hat{S}$ represents the output construction sequence, $N_s$ denotes the length of the construction sequence, $\hat{S}_c$ is the output CAD command, $\hat{S}_p$ represents the output CAD command parameters, and $\Theta$ refers to the learnable model parameters.  

The decoder output is passed through two separate linear layers, which are used to predict the CAD commands and their corresponding parameters, respectively. The first linear layer, $W_1 \in \mathbb{R}^{768 \times 6}$, predicts six CAD commands ($\hat{S}_c$). The second linear layer $W_2 \in \mathbb{R}^{768 \times 4096}$ predicts the 16 command parameters ($\hat{s}_p$). To facilitate this, $W_2$ is reshaped into $\mathbb{R}^{768 \times 16 \times 256}$ to predict both the 16 command parameters and their respective values within the range of 0 to 255. The learnable parameters are optimized using a loss function, as defined by the following equation, to accurately predict the CAD commands and their corresponding parameters. 
\begin{equation}
\mathcal{L} = \sum_{i=1}^{N_s} l(\hat{S}_{c_i}, S_{c_i}) + \sum_{i=1}^{N_s} \sum_{j=1}^{N_p} l(\hat{S}_{p_{i,j}}, S_{p_{i,j}}),
\end{equation}
where $\ell(\cdot, \cdot)$ denotes the Cross-Entropy loss function, $S_{c_i}$ represents the predicted CAD command and $S_{p_{i,j}}$ corresponds to the predicted CAD command parameter.


\section{Experimental Setup and Details}
\label{sec:Experiments}
\subsection{Dataset}
For the experiments, we constructed the GuideCAD dataset, which comprises paired construction sequences, text prompts, and 3D CAD images, based on the DeepCAD \cite{deepcad} dataset. The text prompts were tokenized using the GPT-2 tokenizer, with a sequence length of $N_t=512$ tokens. Consistent with DeepCAD \cite{deepcad}, the maximum command length was set to $N_s=60$. The dataset consists of pairs of text prompts and 3D CAD images, which were created by rotating the 3D CAD model along the XYZ axes to generate seven distinct views. For each training iteration, one image was randomly selected, resized to $224\times224$ pixels, and paired with the corresponding text prompt. During the validation and test phases, GuideCAD predicts the construction sequence from the input data, which is composed of pair of a single image, rotated by ($\frac{\pi}{6}$ $\frac{\pi}{6}$, $\frac{\pi}{6}$) on the XYZ axes, along with its corresponding text prompt.

\subsection{Implementation details}
GuideCAD employs the pretrained GPT-2 \cite{gpt2} as the language model and the pretrained CLIP \cite{clip} for the image encoder, with the pretrained parameters frozen during training. The Mapping Network ($f_m$) consists of two transformer encoders, each with 768-dimensional embeddings, and $p_l$ set to 5. In the Adaptive Pooling Layer, $E_a$  is set to 32, and a Transformer encoder with a 768-dimensional embedding is employed. The decoder is made up of four Transformer Decoder blocks, where each block includes a 768-dimensional embedding, a 512-dimensional feedforward embedding, and 8 multi-head attention layers, with a dropout rate of 0.1. Training was conducted on the 3 Tesla V100-PCIE-32GB GPUs with a batch size of 512, applying  half-precision \cite{fp16} and using gradient accumulation over four steps for 60 epochs. The AdamW optimizer \cite{adamw} was employed with a learning rate of $1e-3$, and the CosineAnnealingLR scheduler was applied, setting the minimum learning rate ($\eta_{min}$) to $1e-4$ and $T_{max}$ to 10.

\subsection{Fine-tuning Guideline}
For fine-tuning GuideCAD, we updated all parameters except for the image encoder. The fine-tuned model was trained for 100 epochs with a batch size of 128, and gradient accumulation was applied four times. The AdamW optimizer \cite{adamw} was used with a learning rate of $1e-4$. Furthermore, a CosineAnnealingLR scheduler was implemented, setting the minimum learning rate to $1e-5$ and $T_{max}$ to 10. All other hyperparameters were maintained as specified in the Implementation Details section.

\subsection{Metrics}
We evaluate the performance of 3D CAD model generation from three main aspects. First, we use the F1 score and Accuracy to assess how well the predicted construction sequence matches the ground truth (GT). Second, to evaluate the quality of the generated 3D CAD models, we compute the Chamfer Distance (CD), Coverage (COV), Minimum Matching Distance (MMD), and Jensen–Shannon Divergence (JSD). Third, we calculate the proportion of failed 3D CAD model generations using the Invalidity Ratio (IR).

\subsubsection{Evaluation Metrics for Construction Sequence}
To evaluate the performance of 3D CAD model generation, we compared the F1-score and the accuracy metric used in DeepCAD \cite{deepcad}. The comparison was conducted by examining the commands and command parameters between the ground truth construction sequence ($S$) and the construction sequence ($\hat{S}$) predicted by the model.

\textbf{F1 score}. To compare $\hat{S}$ and $S$, Precision and Recall were calculated by evaluating the values of True Positives (TP), False Positives (FP), and False Negatives (FN), as presented in the following equation.
\begin{align}
TP = \sum_{i=1}^{N_s} \left[ s_{c_i} = \hat{s}_{c_i} \right], \\
FP = \sum_{i=1}^{N_s} \left[ s_{c_i} \neq \hat{s}_{c_i}, \hat{s}_{c_i} \neq \emptyset \right], \\
FN = \sum_{i=1}^{N_s} \left[ s_{c_i} \neq \hat{s}_{c_i}, s_{c_i} \neq \emptyset \right],
\end{align}
Here, TP represents the case where the command values between $S$ and $\hat{S}$ match exactly, FP refers to cases where the predicted command values in $\hat{S}$ are incorrect when compared to $S$, and FN indicates cases where a command exists in $S$, but the corresponding value in $\hat{S}$ is either incorrectly predicted or missing.

Recall and F1-score were computed for each of the four commands (Line, Arc, Circle, Extrusion) based on the respective TP, FN, and FP values, as follows:
\begin{align}
\text{Precision} = \frac{TP}{TP + FP}, \\ 
\text{Recall} = \frac{TP}{TP + FN}, \\
\text{F1} = 2 \times \frac{\text{Precision} \times \text{Recall}}{\text{Precision} + \text{Recall}}.
\end{align}

\textbf{Command Accuracy}. Each of the four commands was compared between 
$S$ and $\hat{S}$ respectively, and the accuracy was calculated according to the following equation:
\begin{equation}
\text{ACC}_{(S_c)} = \frac{1}{N_s} \sum_{i=1}^{N_s} \left[ s_{(c_i)} = \hat{s}_{(c_i)} \right].
\end{equation}

\textbf{Parameter Accuracy}. The accuracy of the predicted parameter values for the four commands in $\hat{S}$, corresponding to each command, was calculated by comparing the parameter values between $S$ and $\hat{S}$, as follows:
\begin{equation}
\text{ACC}_{(S_p)} = \frac{1}{M} \sum_{i=1}^{|N_s|} \sum_{j=1}^{|s_p|} \mathbb{I} \left[ |s_{p_{(i,j)}} = \hat{s}_{p_{(i,j)}} | < \eta \right] \mathbb{I} \left[ s_{c_i} = \hat{s}_{c_i} \right],
\end{equation} 
where $\mathbb{I}[\cdot]$ is the indicator function (0 or 1), $M = \sum_{i=1}^{N_s} \mathbb{I}[s_{(c_i)} = \hat{s}_{(c_i)}] |s_{(p_i)}|$ represents the total number of correctly predicted commands, and $\eta$ is the tolerance threshold for parameter quantization, set to 3 as in DeepCAD \cite{deepcad}, allowing for minor discrepancies between predicted and actual parameter values.

\subsubsection{Evaluation Metrics for 3D CAD Model Quality}
We evaluated the quality of 3D CAD models by calculating the Coverage (COV), Minimum Matching Distance (MMD), and Jensen-Shannon Divergence (JSD) metrics, comparing the ground truth with the output generated by the baseline model. To address this, we sampled 1000 point clouds from both the ground truth and each generated 3D CAD model, defining the sampled sets as $G$ and $\hat{G}$, respectively. This approach follows the method used in DeepCAD \cite{deepcad}, where similar metrics were applied to evaluate the quality of the generated CAD model.

\textbf{Coverage (COV).} To measure the diversity of 3D CAD model shapes between \( G \) and \( \hat{G} \), we calculate Coverage (COV), which calculates the proportion of shapes in the reference set $G$ that are covered by at least one shape in the generated set $\hat{G}$, as defined by the following equation:
\begin{equation}
\text{COV}(G, \hat{G}) = \frac{|\{ \arg \min_{u \in G} \text{CD}(u,v) \mid v \in \hat{G} \}|}{|G|},
\end{equation}
where $\text{CD}(u,v)$ denotes the chamfer distance between two point clouds $u$ and $v$.

\textbf{Minimum Matching Distance (MMD).} MMD evaluates the fidelity of generated shapes by computing the Chamfer distance between each shape in the reference set $G$ and its nearest neighbor in the generated set $\hat{G}$. The MMD is calculated as the average of these minimum distances, as defined by the following equation:
\begin{equation}
\text{MMD}(G, \hat{G}) = \frac{1}{|G|} \sum_{u \in G} \left( \min_{v \in \hat{G}} \text{CD}(u,v) \right),
\end{equation}
where $\text{CD}(u,v)$ denotes the chamfer distance between two point clouds $u$ and $v$.

\textbf{Jensen-Shannon Divergence (JSD).} We measure the similarity between $G$ and $\hat{G}$ by calculating the Jensen-Shannon Divergence, which is computed by repeating the calculation three times to reduce sampling bias, defined as follows:
\begin{equation}
\text{JSD}(P_{\text{G}}, P_{\text{model}}) = \frac{1}{2} D_{\text{KL}}(P_{\text{G}} \| P_{\text{avg}}) + \frac{1}{2} D_{\text{KL}}(P_{\text{model}} \| P_{\text{avg}}),
\end{equation}
where \( P_{\text{avg}} = \frac{1}{2}(P_{\text{G}} + P_{\text{model}}) \), \( D_{\text{KL}} \) denotes the Kullback-Leibler divergence, and \( P_{\text{G}} \) and \( P_{\text{model}} \) are the marginal distributions of points in the $G$ and $\hat{G}$, respectively. To approximate the point cloud distributions, we approximated the distribution of each point cloud by randomly selecting 1,000 points from set $G$ and 3,000 points from set $\hat{G}$, and then mapping these sampled points into voxel grids discretized to $28^3$. 

\subsection{Baseline}
We compare the performance with the following baseline approaches:
\begin{itemize} 
\item {\textbf{DeepCAD \cite{deepcad}.}} As DeepCAD \cite{deepcad} does not support multi-modal inputs such as image-text data, we trained the DeepCAD decoder using $z_{gpt}$ as the ground truth and conditional input to generate 3D CAD models from text prompts and images. During inference, the predicted $z_{gpt}$ from GPT-2, along with the constant embedding, is passed to the decoder to generate the construction sequence. 

\item {\textbf{OpenECAD \cite{OpenECAD}.}} To compare with GuideCAD, we evaluated the OpenECAD-CLIP-0.55B model, trained according to the guidelines in the corresponding paper. It generates Python script based on pythonOCC, which is then converted into a .step file and used to generate the 3D CAD model. 

\item {\textbf{Cross-modal alignment based SOTA VLMs.}} To validate the effectiveness of our prefix-based embedding alignment strategy, we compare it with state-of-the-art VLMs designed for cross-modal alignment (\textit{e.g.,} LlaVA-NeXT \cite{liu2024llavanext}, Qwen2.5-VL \cite{Qwen2.5-VL}). To match the decoder’s input dimension, the average embedding from the final layer of the VLM is transformed via a linear projection. This projected embedding is then used as a conditional input and treated as the ground truth for training the decoder. At inference, predicted average embedding from the final layer of VLM, combined with the constant embedding, was provided as input to the decoder to predict the construction sequence. 

\end{itemize}

\section{Experimental Results}

\begin{figure}[!htbp]
\centering
\includegraphics[width=0.9\textwidth, height=0.92\textheight]{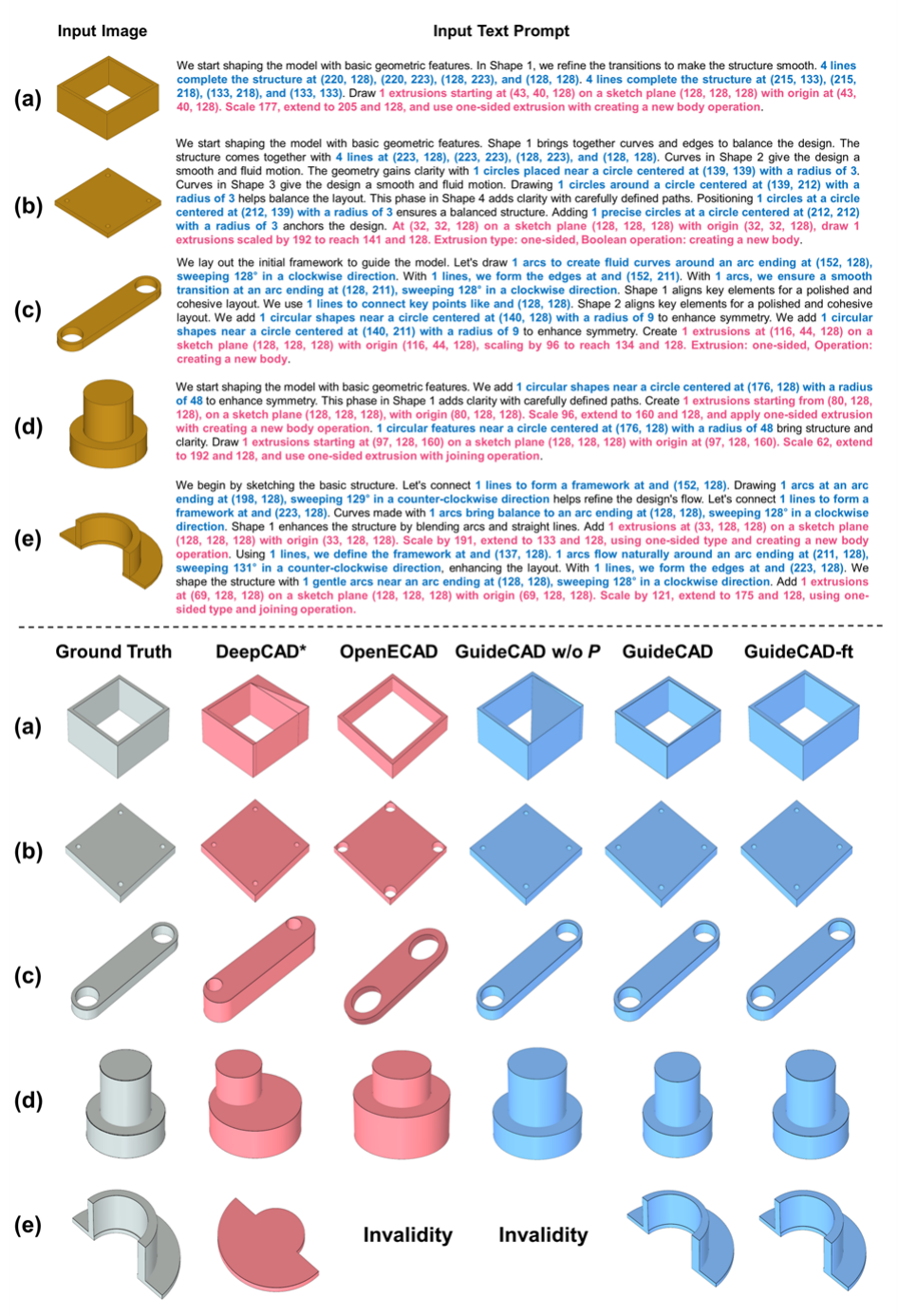}
\caption{Qualitative comparison of reconstructed CAD models from baseline models on the test dataset. The blue text in the input prompt represents the \textit{Sketch} commands (Line, Circle, Arc), while the pink text indicates the \textit{Extrusion} commands. The "ft" suffix represents fine-tuning, while an asterisk (*) marks models adapted to our training and inference process.}
\label{fig:figure4}
\end{figure}

\begin{table}[t]
\caption{
Summarizes the performance comparison of baseline models on the GuideCAD dataset. The suffix ``ft'' denotes a fine-tuned model, while an asterisk (*) indicates models adapted to our training and inference process.
}
\label{tab:table1}
\centering
\resizebox{\textwidth}{!}{%
\begin{tabular}{l cccc ccccc c}
\hline
\multirow{2}{*}{\textbf{Model}} &
  \multicolumn{4}{c}{\textbf{F1 score ($\times10^2$) $\uparrow$}} &
  \multicolumn{5}{c}{\textbf{Quality of 3D Shape}} &
  \multirow{2}{*}{\textbf{IR ($\times10^2$) $\downarrow$}} \\ \cline{2-10} 
 &
  \textbf{Line} &
  \textbf{Arc} &
  \textbf{Circle} &
  \textbf{Extrusion} &
  \textbf{Median CD ($\times10^3$) $\downarrow$} &
  \textbf{Mean CD ($\times10^3$) $\downarrow$} &
  \textbf{COV ($\times10^2$) $\uparrow$} &
  \textbf{MMD ($\times10^2$) $\downarrow$} &
  \textbf{JSD ($\times10^2$) $\downarrow$} & 
  \\ \hline
DeepCAD$^*$ \cite{deepcad} &
  89.96 & 60.59 & 84.35 & 81.10 &
  5.75 & 83.69 & 79.23 & 1.41 & 1.87 &
  16.11 \\
OpenECAD \cite{OpenECAD} &
  \multicolumn{4}{c}{-} &
  9.61 & 52.85 & 51.06 & 2.34 & 2.55 &
  12.22 \\
\textbf{GuideCAD w/o $P$ (Ours)} &
  \textbf{93.67} & \textbf{73.49} & \textbf{90.67} & \textbf{88.50} &
  \textbf{2.35} & \textbf{24.58} & \textbf{81.07} & \textbf{1.26} & \textbf{1.57} &
  \textbf{8.97} \\
\textbf{GuideCAD (Ours)} &
  \textbf{94.52} & \textbf{78.11} & \textbf{91.79} & \textbf{90.57} &
  \textbf{2.06} & \textbf{17.66} & \textbf{83.43} & \textbf{1.16} & \textbf{1.50} &
  \textbf{7.62} \\
\textbf{GuideCAD-ft (Ours)} &
  \textbf{96.89} & \textbf{86.09} & \textbf{94.87} & \textbf{94.12} &
  \textbf{1.77} & \textbf{12.58} & \textbf{83.93} & \textbf{1.14} & \textbf{1.49} &
  \textbf{5.27} \\ \hline
\end{tabular}%
}
\end{table}

\begin{table}[t]
\caption{Performance comparison of four embedding methods such as Cross-modal alignment, Summation, Concatenate, and Prefix.}
\label{tab:table2}
\centering
\resizebox{\textwidth}{!}{%
\begin{tabular}{l l cccc ccccc c}
\hline
\multirow{2}{*}{\textbf{Model}} &
  \multirow{2}{*}{\textbf{Method}} &
  \multicolumn{4}{c}{\textbf{F1 score ($\times10^2$) $\uparrow$}} &
  \multicolumn{5}{c}{\textbf{Quality of 3D Shape}} &
  \multirow{2}{*}{\textbf{IR ($\times10^2$) $\downarrow$}} \\ \cline{3-11}
 & &
  \textbf{Line} &
  \textbf{Arc} &
  \textbf{Circle} &
  \textbf{Extrusion} &
  \textbf{Median CD ($\times10^3$) $\downarrow$} &
  \textbf{Mean CD ($\times10^3$) $\downarrow$} &
  \textbf{COV ($\times10^2$) $\uparrow$} &
  \textbf{MMD ($\times10^2$) $\downarrow$} &
  \textbf{JSD ($\times10^2$) $\downarrow$} &
  \\ \hline
LlaVA-NeXT \cite{liu2024llavanext} &
  Cross-modal alignment &
  90.11 & 67.02 & 85.40 & 80.38 &
  15.20 & 106.12 & 73.70 & 1.64 & 2.32 &
  16.60 \\
Qwen2.5-VL \cite{Qwen2.5-VL} &
  Cross-modal alignment &
  91.28 & 69.05 & 86.11 & 82.90 &
  11.36 & 91.43 & 76.57 & 1.54 & 2.15 &
  13.40 \\
GuideCAD &
  Summation &
  93.34 & 72.60 & 90.29 & 88.02 &
  2.30 & 25.67 & 82.07 & 1.22 & 1.50 &
  9.82 \\
GuideCAD &
  Concatenation &
  93.24 & 72.42 & 89.86 & 87.85 &
  2.25 & 23.17 & 82.97 & 1.20 & 1.52 &
  9.85 \\
\textbf{GuideCAD} &
  \textbf{Prefix} &
  \textbf{94.52} & \textbf{78.11} & \textbf{91.79} & \textbf{90.57} &
  \textbf{2.06} & \textbf{17.66} & \textbf{83.43} & \textbf{1.16} & \textbf{1.49} &
  \textbf{7.62} \\ \hline
\end{tabular}%
}
\end{table}

\begin{figure}[!htbp]
\centering
\includegraphics[width=0.6\linewidth,height=0.15\textheight]{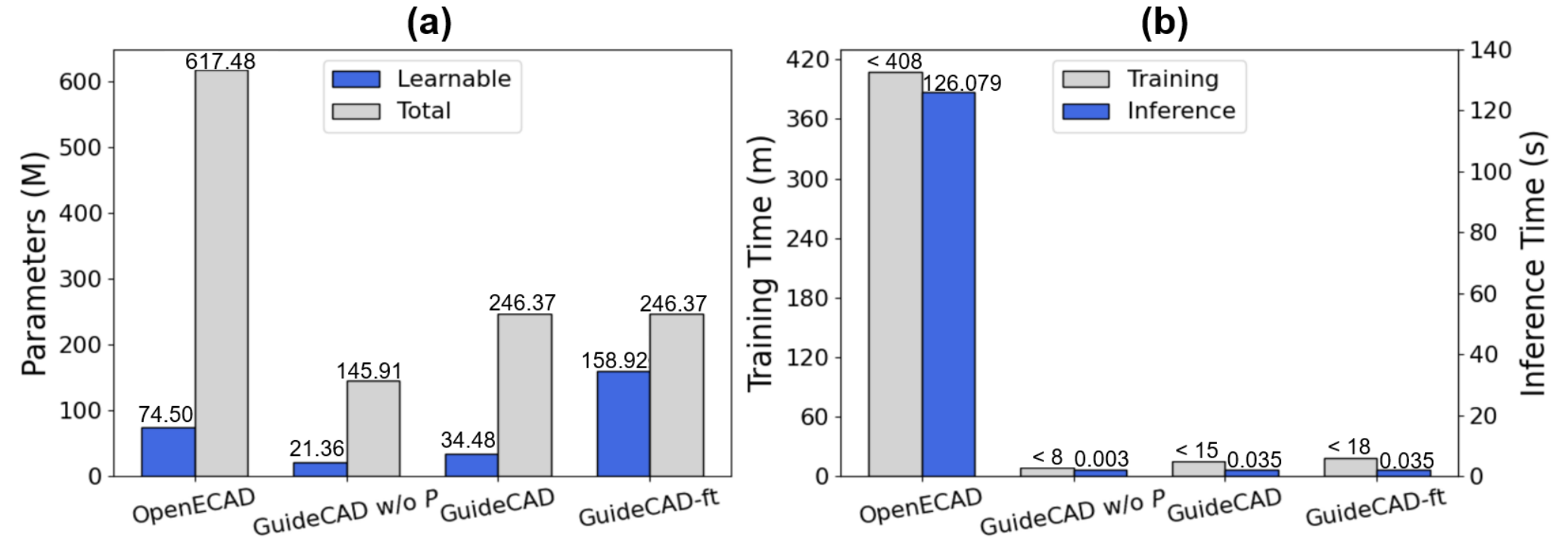}
\caption{Evaluation of the baseline in terms of the number of learnable parameters and computation time. (a) Comparison based on the number of learnable parameters (b) Analysis of training and inference time. The training time is measured for one epoch on the training dataset, and inference time is the average time over 1,000 samples randomly selected from the test dataset.}
\label{fig:figure5}
\end{figure}

\begin{figure}[!htbp]
\centering
\includegraphics[width=0.9\linewidth]{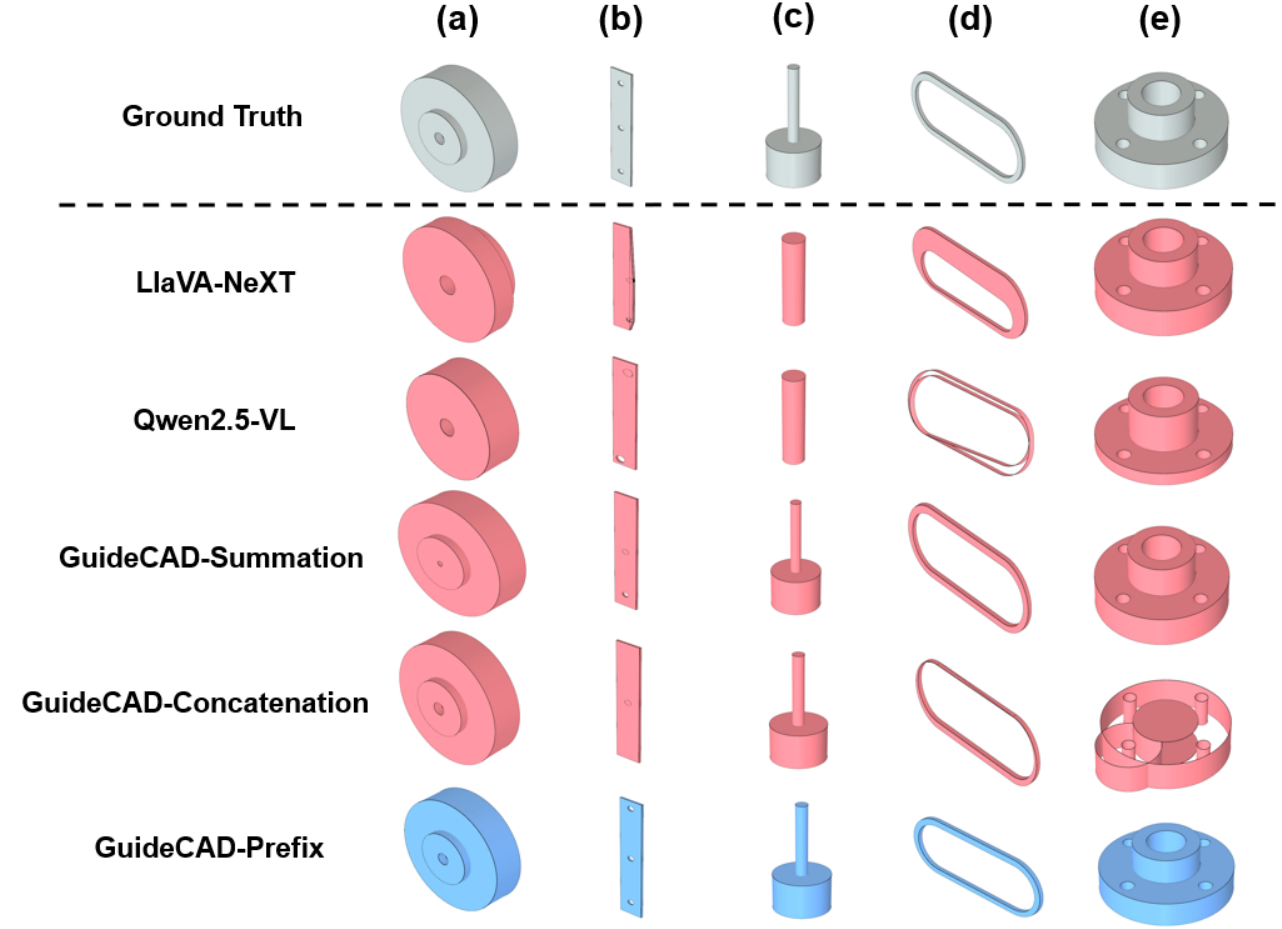}
\caption{Comparison of CAD model generation using different embedding methods: LlaVA-NeXT, Qwen2.5-VL, Summation, Concatenation, and Prefix.}
\label{fig:figure6}
\end{figure}

\begin{figure}[!htbp]
\centering
\includegraphics[width=\linewidth, height=0.14\textheight]{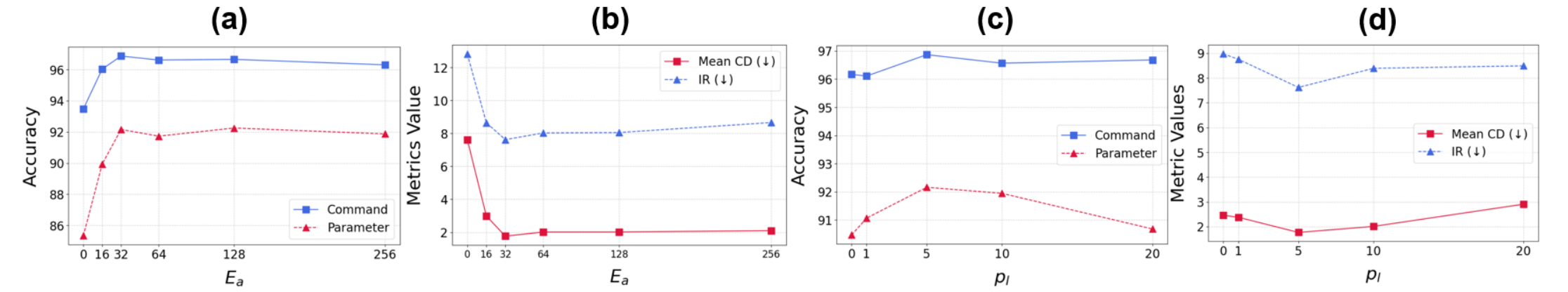}
\caption{Performance analysis based on variations in $E_a$ and prefix length ($p_l$). (a) Comparison of command and command parameter accuracy with changes in $E_a$. (b) Comparison of Mean CD and IR with changes in $E_a$. (c) Comparison of command and command parameter accuracy with changes in $p_l$. (d) Comparison of Mean CD and IR with changes in $p_l$.}
\label{fig:figure7}
\end{figure}

\subsection{Qualitative evaluation of baselines}
Figure \ref{fig:figure4} compares the 3D CAD model generation performance with baseline models, using text prompts and 3D CAD images as inputs. DeepCAD \cite{deepcad} generates 3D CAD models that are similar to the ground truth by predicting the construction sequence, but it fails to accurately predict parameters such as the position of the Line, the center and radius of the Circle, and the scale and type of Extrusion. In contrast, as shown in Figure \ref{fig:figure4} (a-d), OpenECAD \cite{OpenECAD} generates CAD models that are generally similar to the ground truth. However, for more complex shapes, the generated Python code becomes longer, leading to a decrease in the accuracy of details in the CAD model. Specifically, as shown in Figure \ref{fig:figure4} (e), OpenECAD produces invalid outputs for more complex shapes. In contrast, GuideCAD efficiently generates 3D CAD models that closely match the ground truth by learning only a small number of additional parameters, thereby outperforming the baselines.

\subsection{Comparison of Baselines in Construction Sequence}
Table \ref{tab:table1} quantitatively compares the quality of the construction sequences generated from the test dataset. GuideCAD outperforms DeepCAD \cite{deepcad} in terms of F1 score, demonstrating significant improvements: 28.92\% and 11.68\% for the Arc and Extrusion commands, respectively, and 8.82\% and 5.06\% for the Circle and Line commands, respectively. As OpenECAD generates Python code directly rather than predicting the construction sequence, it is difficult to compare the command prediction accuracy using the F1 score. Therefore, the evaluation was based on the quality of the generated 3D CAD models. As a result, GuideCAD achieves better results than both DeepCAD and OpenECAD across all metrics, most notably in the Median CD, with relative improvements of 63.74\% and 78.59\%, respectively. These results show that predicting a construction sequence generates more consistent and accurate 3D CAD models than directly generating Python code. 

\subsection{Analysis of Model Complexity and Computation Cost}
To compare the number of learnable parameters and computational cost, we present the results in Figure \ref{fig:figure5}, in comparison with state-of-the-art vision-language models. GuideCAD, which learns $P$ via a mapping network, requires more training time and a higher number of learnable parameters than GuideCAD w/o $P$. However, they exhibit similar inference times, indicating negligible runtime difference in generating 3D CAD models. Although OpenECAD \cite{OpenECAD} employs approximately twice the number of learnable parameters, GuideCAD generates high-fidelity 3D CAD models that better reflect the intended design, as shown in Figure~\ref{fig:figure4}. In addition, GuideCAD requires significantly less training and inference time compared to OpenECAD. These results show that GuideCAD is more efficient than existing vision-language models in generating 3D CAD models.

\subsection{Evaluation of Cross-modal Embedding Strategies}
We conducted experiments to evaluate various cross-modal embedding strategies. Table \ref{tab:table2} compares the performance of four approaches (Summation, Concatenate, Vision-Language Projection, and Prefix) in generating 3D CAD models. The summation and concatenation approaches integrate image and text embeddings without relying on the prefix embedding (P) generated by a mapping network. In the summation approach, the image embedding is transformed via a linear layer to have the same dimensionality as the text embedding and is combined with it through element-wise addition before being fed into GPT-2 as input. In the concatenation approach, the image embedding is reshaped to align with the sequence length dimension of the text embeddings, after which it is concatenated with the text token sequence and jointly used as input to GPT-2 during training. The results show that the prefix-based approach (\textit{i.e.,} ours) outperforms the other methods in terms of all metrics. Furthermore, Figure \ref{fig:figure6} demonstrates that GuideCAD-Prefix generates high-quality 3D CAD models that closely resemble the ground truth, achieving better performance than other embedding methods. These results suggest that prefix-based embedding is more effective than other methods in generating 3D CAD models. 

\subsection{Comparison with GPT-2 Fine-Tuned GuideCAD}
In this experiment, we compared the 3D CAD model generation performance and efficiency of GuideCAD and Fine-Tuned GuideCAD (GuideCAD-ft). As presented in Table \ref{tab:table1} and Figure \ref{fig:figure5}, the GuideCAD-ft, which fine-tunes all parameters except the pretrained image encoder based on the fine-tuning guidelines, uses approximately 4.61 times more learnable parameters than GuideCAD and requires approximately 16.67\% more training time per epoch. Although GuideCAD-ft shows improved F1 scores, better 3D CAD model quality, and a reduced Invalidity Ratio (IR), the differences between GuideCAD and GuideCAD-ft were not significant. Figure \ref{fig:figure4} illustrates that the 3D CAD models generated by GuideCAD and GuideCAD-ft show no significant difference, with both being similar to the ground truth. These results demonstrate that when training costs and computational resources are limited, learning a small number of parameters through prefix embedding offers an efficient alternative. On the other hand, fine-tuning depends on the trade-off between accuracy and computational cost, and it is preferable when aiming to generate a high-quality CAD model.

\subsection{Ablation Study}
\subsubsection{Effectiveness of Prefix-Assisted Input}
To analyze the impact of image prefix embedding ($P$) on 3D CAD model generation, we compare GuideCAD w/o $P$ to GuideCAD in Table \ref{tab:table1}. GuideCAD shows superior performance to GuideCAD w/o $P$ by predicting construction sequences more accurately and generating more valid CAD models. Specifically, the prediction accuracy for the Arc and Extrusion commands increased by 6.29\% and 2.34\%, respectively. In terms of 3D shape quality, the Median CD and Mean CD values both decreased, while the COV, MMD, and JSD metrics all improved. These results are presented in Figure \ref{fig:figure4}, where GuideCAD generated 3D CAD models more similar to the ground truth, while GuideCAD w/o $P$ produced models that were either incorrect or invalid. To summarize, including image information as a prefix along with text prompts enhances the generation of high-quality 3D CAD models. 

\subsubsection{Impact of Adaptive Pooling Layer}
In this experiment, we evaluated the performance of the Adaptive Pooling Layer (APL) by varying the $E_a$ value. Figure \ref{fig:figure7} (a) shows the accuracy of the predicted construction sequence, while Figure \ref{fig:figure7} (b) presents the Mean CD and invalidity ratio. When $E_a$=0, APL is not used, resulting in the highest invalidity ratio because commands and parameters show the lowest accuracy, and low quality of generated 3D shape.  As $E_a$ increases, accuracy improves steadily, reaching its peak at $E_a$=32, where the best performance is achieved in terms of Accuracy, Mean CD, and IR. Beyond this point, further increases in $E_a$ lead to degraded performance, indicating that excessively large embedding sizes introduce inefficiencies. This finding suggests that a too-small embedding size fails to learn enough relevant information, while a too-large size adds complexity, reducing training efficiency.

\subsubsection{Performance Analysis of Prefix Length}
To evaluate the impact of increasing prefix length ($p_l$) on performance, Figure \ref{fig:figure7} (c) shows the accuracy of the commands and parameters, while Figure \ref{fig:figure7} (d) shows the Mean CD and IR. To generate valid 3D CAD models, accurate prediction of the construction sequence leads to a lower invalidity ratio, while a low Mean CD reflects similarity to the ground truth. Although an increase in $p_l$ was expected to improve overall performance, no significant improvement was observed, with Accuracy, Mean CD, and IR values peaking at $p_l$=5 and then decreasing. These results indicate that the mapping network does not perfectly transfer information from the pretrained image encoder, thereby limiting its ability to generalize 3D information from a single image.

\section{Limitations and Future Work}
\label{sec:Limitations and Future Work}
The GuideCAD dataset is designed to enable the generation of text prompts based on detailed values expressed in the corresponding construction sequence. Fixed templates restrict the expressive capacity of text prompts and fail to utilize visual information. To overcome this, our method augments text prompts with visual information using vision-language models \cite{liu2024llavanext, Qwen2.5-VL}. Although we adopted GPT-2 and CLIP as lightweight models for CAD model generation, the performance could potentially be improved by utilizing more powerful visual encoders or more recent large language models. Furthermore, incorporating consistency information extracted from multi-view images may serve to further enhance the performance of the mapping network. An alternative approach leverages an autoregressive language model to directly generate 3D CAD models as Python script without the need for a decoder \cite{llamamesh,rukhovich2024cadrecode}, thereby reducing model complexity and improving training efficiency.

\section{Conclusion}
\label{sec:conclusion}
In this study, we propose GuideCAD, a novel framework for efficiently learning from image-text multi-modal data to generate 3D CAD models. A key component of our approach is the Mapping Network, which transforms image embeddings into prefix embeddings. It allows the language model to effectively incorporate visual information into the textual information. Our experimental results show that GuideCAD achieves competitive performance in generating 3D CAD models, while using approximately four times fewer parameters and training more efficiently than fine-tuning approaches. Furthermore, we demonstrate that high-fidelity 3D CAD models can be generated by leveraging semantically rich embeddings from pretrained models, without requiring full fine-tuning.

\section*{Acknowledgements}
This work was supported in part by the National Research Foundation of Korea (NRF) grant funded by the Korea government(MSIT) (RS-2024-00341307) and was supported in part by the IITP(Institute of Information \& Coummunications Technology Planning \& Evaluation)-ICAN(ICT Challenge and Advanced Network of HRD) grant funded by the Korea government(Ministry of Science and ICT)(IITP-2025-RS-2024-00437024). This work was also supported by the National Research Foundation of Korea(NRF) grant funded by the Korea government(MSIT) (RS-2024-00346364).

\bibliographystyle{unsrtnat}
\bibliography{references}

\end{document}